\def\year{2022}\relax
\documentclass[letterpaper]{article} 
\usepackage{aaai22}  
\usepackage{times}  
\usepackage{helvet}  
\usepackage{courier}  
\usepackage[hyphens]{url}  
\usepackage{graphicx} 
\usepackage{natbib}  
\usepackage{caption} 
\DeclareCaptionStyle{ruled}{labelfont=normalfont,labelsep=colon,strut=off} 
\usepackage{times}  
\usepackage{helvet}
\usepackage{courier} 
\usepackage{graphicx}
\usepackage{caption}
\usepackage{amsmath}
\usepackage{amsthm}
\usepackage{times}
\usepackage{pdfpages}
\usepackage{multirow}
\usepackage{multicol}
\usepackage{bm}
\usepackage{subfigure}
\usepackage{float}
\usepackage{amssymb}
\usepackage{booktabs}
\usepackage{newfloat}
\usepackage{listings}
\floatstyle{ruled}
\newfloat{listing}{tb}{lst}{}
\floatname{listing}{Listing}
\title{Learning Task-aware Robust Deep Learning Systems}
\author{
   	$^{1,2}$Keji Han,
   	$^{1,2}$Yun Li\thanks{Corresponding Author},
   	$^{1,2}$Xianzhong Long,
   	$^{1,2}$Yao Ge,
    \\
}
\affiliations{
	 \textsuperscript{\rm 1}Nanjing University of Posts and Telecommunications\\
  \textsuperscript{\rm 2}Jiangsu Key Laboratory of Big Data Security \& Intelligent Processing

   liyun@njupt.edu.cn

}

\usepackage{bibentry}

\begin{document}

\maketitle
\renewcommand{\thefootnote}{}
\nonumber\footnote{For Preview}
\begin{abstract}
Many works demonstrate that deep learning system is vulnerable to adversarial attack. A deep learning system consists of two parts: the deep learning task and the deep model. Nowadays, most existing works investigate the impact of the deep model on robustness of deep learning systems, ignoring the impact of the learning task. In this paper, we adopt the binary and interval label encoding strategy to redefine the classification task and design corresponding loss to improve robustness of the deep learning system. Our method can be viewed as improving the robustness of deep learning systems from both the learning task and deep model. Experimental results demonstrate that our learning task-aware method is much more robust than traditional classification while retaining the accuracy.
\end{abstract}
\section{Introduction}
Deep neural networks have been applied to many learning tasks, such as image recognition, speech recognition, and machine translation \cite{deepl}. With increasing real-world applications, the robustness of deep neural networks arouses increasing attention from both academia and industry. Deep neural networks are demonstrated to be vulnerable to the adversarial example \cite{adve}. The adversarial example is crafted by adding imperceptible adversarial perturbation to the original legitimate example. In essence, the existence of adversarial examples is rooted in the difference between human intelligence and machine intelligence \cite{kemlpda}.\\\\
There are many adversarial attack methods proposed to explore the vulnerability of deep learning systems. According to the phase that adversarial attack happens, adversarial attack methods fall into two categories: poisoning attack and evasion attack \cite{attackanddefense}. In this paper, we focus on the evasion attack. According to manner to craft adversarial examples, existing evasion attack methods can be divided into two categories, namely single-step attack and multi-step attack. A single-step attack explores adversarial perturbation in one step or directly maps the original example as an adversarial example. Multi-step attack explores adversarial perturbation iteratively. For the single-step attack, FGSM \cite{fgsm} crafts the adversarial example with the gradient sign; AdvGAN \cite{advgan} and ATN \cite{atn} directly map the legitimate example as an adversarial example. As to the multi-step attack, PGD \cite{pgd} crafts the adversarial example in an iterative way; CW \cite {cw} and Deepfool \cite{deepfool} formulate the attack as an optimization problem then solve it in an iterative way.\\\\
Existing methods to improve the robustness of deep learning systems focus on even all elements related to the deep model, such as the training data/features, the model architecture, training loss/regularization loss, and parameter-updating strategy. For training data-level methods, feature nullification \cite{fn}, image compress \cite{compression}, and feature squeezing \cite{fs} are demonstrated to be efficient to improve the robustness of the deep learning system. As to the model architecture-level method, the denoising module \cite{featuredenoising}, Euler skip connection \cite{euler}, and additional batch normalization module \cite{mbn} are introduced as additional modules to improve the robustness of the deep model. When it comes to training loss-based methods, adversarial training methods, such as TRADES \cite{trades} and MART \cite{mart}, introduce regularization loss. For the parameter-updating strategy methods, Reluplex \cite{reluplex} updates model parameters with the simplex-like method. Some generative classifiers \cite{genec} update the parameter of deep model with Bayesian Backpropagation (BBP) \cite{bbp}. Moreover, the adversarial example detection methods \cite{magnet, NIC} can keep the adversarial example away from the deep model, which is also demonstrated to be efficient in improving the robustness of the deep learning system.\\\\
Actually, a deep learning system consists of the learning task and a deep model. The learning task provides a formalized definition of the problem, while the deep model implements the learning task. So we can also improve the robustness of the deep learning system by defining robust learning tasks. The traditional classification is so simplistic that the model can be lazy. It just needs to remember instead of learning \cite{rember}. Moreover, in the traditional classification, the accuracy and robustness are at odds \cite{odds}. Above all, the learning task-aware method may be promising to address the adversarial issue for deep learning systems.\\\\
Moreover, there are some works that focus on the learning task to improve the accuracy of the deep learning system. For instance, DeepBE \cite{deepbe} adopts binary encoding labels to improve the accuracy of the deep learning system. In this paper, we explore methods that improve the robustness of the deep learning system from both the learning task and model loss prospects. In detail, we introduce robust binary-label classification with a scaling factor to improve the accuracy and robustness. Moreover, we define the interval-label classification, which marks inputs with predefined nonoverlapping intervals.\\\\
Our contributions can be summarized as follows.
\begin{itemize}
\item We introduce task-aware robust deep learning systems based on binary-label and interval-label classifications;
\item Experimental results demonstrate that our method achieves a lower adversarial transfer rate than traditional classification; 
\item Both analysis and experimental results demonstrate that the accuracy and robustness are no more at odds for the robust binary-label classification;
\item We find that the adversarial training does not constantly improve the robustness of robust binary-label classification, demonstrating that learning task affects the adversarial robustness.
\end{itemize} 
\section{Interval-label and Robust Binary-label classification Systems}
\begin{figure}[htb]
\centering
\includegraphics[width=1\columnwidth]{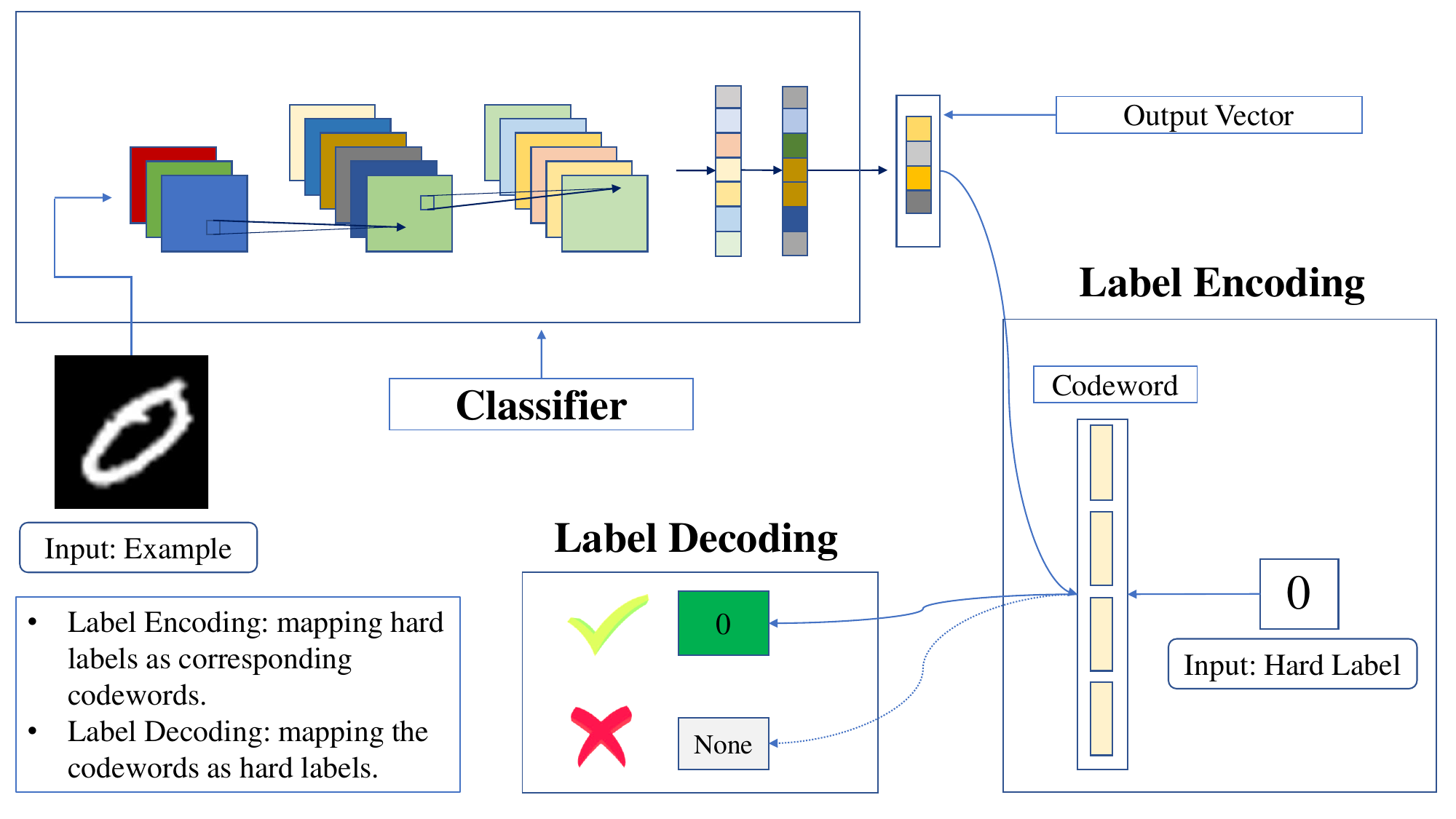}
\caption{The workflow of the learning task-aware robust deep classification system.}\label{workflow}
\end{figure}

Most existing works to improve the robustness of the deep learning system focus on the deep model. As introduced in \cite{kemlpda}, the adversarial example has roots in the difference between machine intelligence and human intelligence. We think that the decision process of the deep learning task may also impact the robustness of the deep learning system. That is another motivation behind proposing robust binary-label and interval-label classification systems.\\\\
A robust classification system consists of three modules: the classifier, label encoding module, and label decoding module, as shown in Fig. \ref{workflow}. The classifier maps inputs into the corresponding output vectors then turns output vectors as codewords. The codewords \cite{ecoc} is the encoding of decimal labels, i.e., hard labels. Label encoding translates hard labels as corresponding codewords, while the label decoding module translates the codewords provided by the classifier as the hard labels, namely the predicted label. If the codeword do not correspond to any predefined hard label, the input will be marked as None, i.e., an abnormal example.
\subsection{Definition}
\textit{Traditional Classification}: the traditional deep classification can be defined as follows: $\forall x\in \mathbb{R}^{D},\mathbf{T}:x\rightarrow \mathbb{R}^{C}$, where $x$ is an example, $D$ is the dimension of $x$. $C$ is the number of categories. \\\\
\textit{Interval-label Classification}: the traditional deep classification can be defined as follows: $\forall x\in \mathbb{R}^{D},\mathbf{T}:x\rightarrow \mathbb{R}^{1}$, where $x$ is an example, $D$ is the dimension of $x$. It predefines some nonoverlapping intervals. The interval that the model output falls is corresponding interval label. If the output does not fall into any predefiend intervals, the input will be marked as an anomaly example.\\\\
\textit{Binary-label Classification}: the binary-label classification can be defined as follows: $\forall x\in \mathbb{R}^{D},\mathbf{T}:x\rightarrow \mathbb{R}^{B}$, where $x$ is an example, $D$ is the dimension of $x$. $B$ is the number of bits in the binary label.\\\\
The decision process of traditional classification is different from the binary-label and interval-label classifications. The traditional classification maximizes the probability corresponding to the ground-truth label. The binary-label classification focuses on each element's sign for the output vector. The interval-label classification tries to decide the interval that the output falls into. Generally speaking, the traditional classification decides the category that the input most likely belongs to, while the binary-label and interval-label classifications have to decide the category that the input exactly belongs to. The following paragraphs will introduce details of robust deep classification systems based on robust binary-label and interval-label classifications.
\subsection{Implementation Details of Interval-label Classification System}
As introduced above, the interval-label classification system consists of three modules: the interval-label classifier, label encoding module, and label decoding module. The interval-label classifier can be a deep neural network whose output is a scalar. The label encoding turns the hard label into an interval label, namely codeword, while the label decoding translates the output of the classifier as a hard label.\\\\
Label Encoding module of interval-label classification system transforms the hard label into an interval label. The map function can be formulated as follows.
\begin{equation}\label{labelmap}
\begin{split}
L(y)=s_{0}+y\cdot(\alpha+\beta);\\ 
U(y)={L}(y)+\beta,\qquad\;\;\;
\end{split}
\end{equation}
where $s_{0}$ is the smallest lower bound of interval labels. $\alpha$ is the length of the gap between two adjacent interval labels, while $ \beta $ is the length of the interval label. $M_{L}(\cdot)$ and $M_{U}(\cdot)$ are lower bound and upper bound map function, respectively. According to Eq. (\ref{labelmap}), for the hard label `3', $s_{0}=0,\alpha=1,\beta=3$, the corresponding interval label is $[12,15]$.\\\\
The label decoding function can be formulated as follows.
\begin{align}
\widetilde{y}=\lfloor\frac{\mathcal{I}(x)-s_{0}}{\alpha+\beta}\rfloor,
\end{align}
where $x$ is the input example. $\mathcal{I}(\cdot)$ is the interval-label classifier. $\lfloor\cdot\rfloor$ is the floor function. If $\mathcal{I}(x)$ does not belong to any interval label, the corresponding input example will viewed as an abnormal example.\\\\
The loss function of the interval-label classification can be formulated as follows.
\begin{equation}\label{int loss}
\begin{split}
\mathcal{L}(B(X,Y^{'});\theta)= \underbrace{\Vert r(\bm{L}(Y)-\mathcal{I}(X))\Vert_{2}^{2}}\limits_{lower\; bound\; loss}+\\
\underbrace{\Vert r(\mathcal{I}(X)-\bm{{U}}(Y)\Vert_{2}^{2}}\limits_{upper\; bound\; loss},\quad
\end{split}
\end{equation}
where $\theta$ is the parameter set of the classifier $\mathcal{I}$. $B(X,Y^{'})$ is a mini batch. $X$ and $Y$ is the examples set and hard label set, and $Y^{'}=[\bm{{L}}(Y), \bm{{U}}(Y)]$ is the interval labels set. $\bm{{L}}(Y)$ and $\bm{{U}}(Y)$ is the lower bound set and upper bound set, respectively. $r(\cdot)$ is the ReLU \cite{relu} activation function. $lower\; bound\; loss$ and $upper\; bound\; loss$ in Eq. \ref{int loss} are the distance between the output and lower and upper bounds, respectively.
\subsection{Implementation Details of Robust Binary-label Classification System}
There are three hyperparameters for interval-label classification, which may limit its generalization. So we also explore another classification for a robust deep classification system, namely robust binary-label classification. As introduced above, a robust binary-label classification (RBC) system involves three modules: the binary-label classifier, the label-encoding module, and the label decoding module. The binary-label classifier is the same as the traditional neural network classifier, except for the output dimension. The label-encoding module converts the traditional decimal label into a binary label, while the label decoding module converts the binary label into a decimal label.\\\\
In this paper, the binary-label classifier is a deep neural network. We set the output dimension of the network as $B$, $2^{B}\geq C$. $C$ is the number of predefined categories. If denoting the robust binary-label classifier as $\mathcal{B}
$, the process to get binary label $\bm{b}$ of the example $x$ can be formulated as follows.
\begin{equation}
\bm{b}_{i}=\left\{
\begin{aligned}
1,\; \mathcal{B}(x)_{i}>0;\\
0,\; \mathcal{B}(x)_{i}\leq 0,
\end{aligned}
\right.\\ 
\end{equation} 
where $i\in 0,1, \cdots, B-1$. $\bm{b}_{i}$ and $\mathcal{B}(x)$ are $i$-th element of the binary label and output of binary-label classifier.\\\\
Label-encoding and label-decoding modules are introduced for conversion between the decimal and binary labels.\\\\
The loss function of the RBC can be formulated as follows.
\begin{equation}\label{loss}
\mathcal{L}(x,\bm{b};\theta)=\|r(S\cdot\mathbf{1}-\mathcal{B}(x)\cdot (2\bm{b}-1))\|^{2}_{2},
\end{equation}
where $x$ and $\bm{b}$ are the input example and its corresponding binary label, respectively. $\theta$ is the parameter set of the binary-label classifier. $r(\cdot)$ is the ReLU activation function. $S$ scaling factor is a hyperparameter to control margins between different categories. $\mathbf{1}$ is the all-ones vector. In detail, $\forall x$ and its binary label $b$, when $S=100$, if $\bm{b}_{i}=0$, the loss in Eq. \ref{loss} will force the $\mathcal{B}(x)_{i}\leq -100$. Otherwise, if $\bm{b}_{i}=1$, the loss in Eq. \ref{loss} will force the $\mathcal{B}(x)_{i}\geq 100$, $i=0,\cdots, B-1$. In other words, loss in Eq. \ref{loss} scales the margin between elements in binary label from $1$ to $2S$.
\section{Experiment}
\subsection{Experiment Settings}
In this paper, two data sets are applied to evaluate the effectiveness of our method, namely MNIST \cite{MNIST}, CIFAR-10, and CIFAR-100 \cite{cifar}. MNIST is a 10-class grayscale hand-written digit image dataset, consisting of 60,000 training examples and 10,000 testing examples. CIFAR-10 is a 10-class color image dataset, consisting of 50,000 training examples and 10,000 testing examples. CIFAR100 is a 100-class color image dataset, consisting of 50,000 training examples and 10,000 testing examples.\\\\
For adversarial robustness evaluation, FGSM and PGD are adopted. FGSM is a classical single-step attack, while PGD is a multi-step attack. The two attacks are adopted for the reason that they can conveniently control the attack intensity. Moreover, since the decision process of the robust binary-label and interval-label classifications are different from the traditional classification, the existing optimization-based attack can not attack the binary-label classification, such as CW and Deepfool. The reason is that adversary can not directly get the probability that the input belongs to a specific category.  
\subsection{Convergence of Robust Binary-label Classification System}
\begin{figure}[htb]
\centering
\subfigure[TRA MNIST]{
\includegraphics[width=0.48\columnwidth]{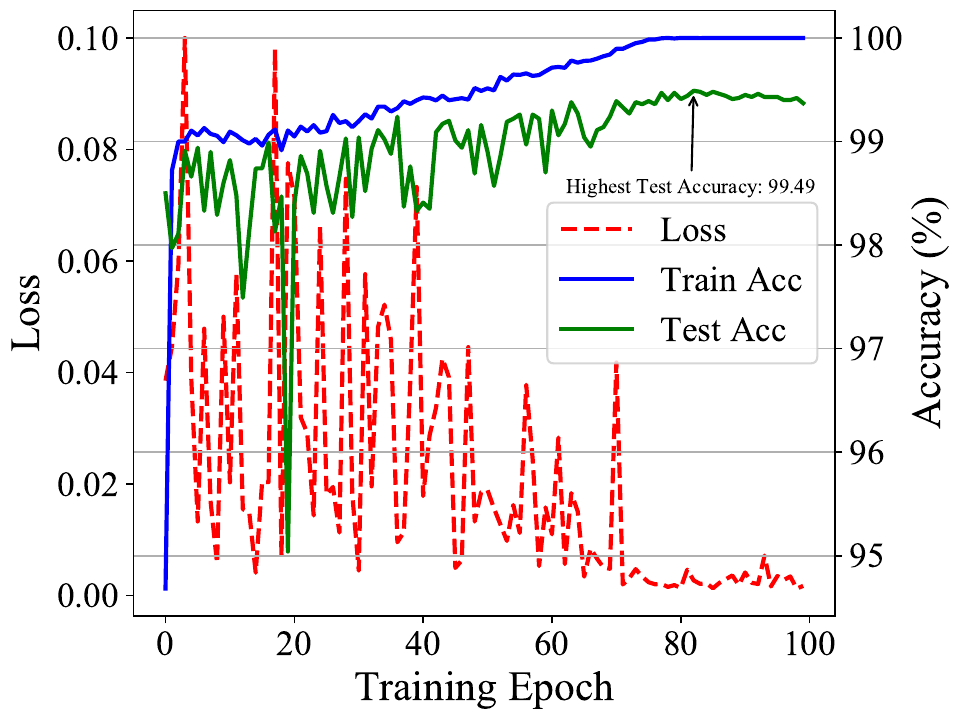}}
\subfigure[RBC MNIST]{
\includegraphics[width=0.48\columnwidth]{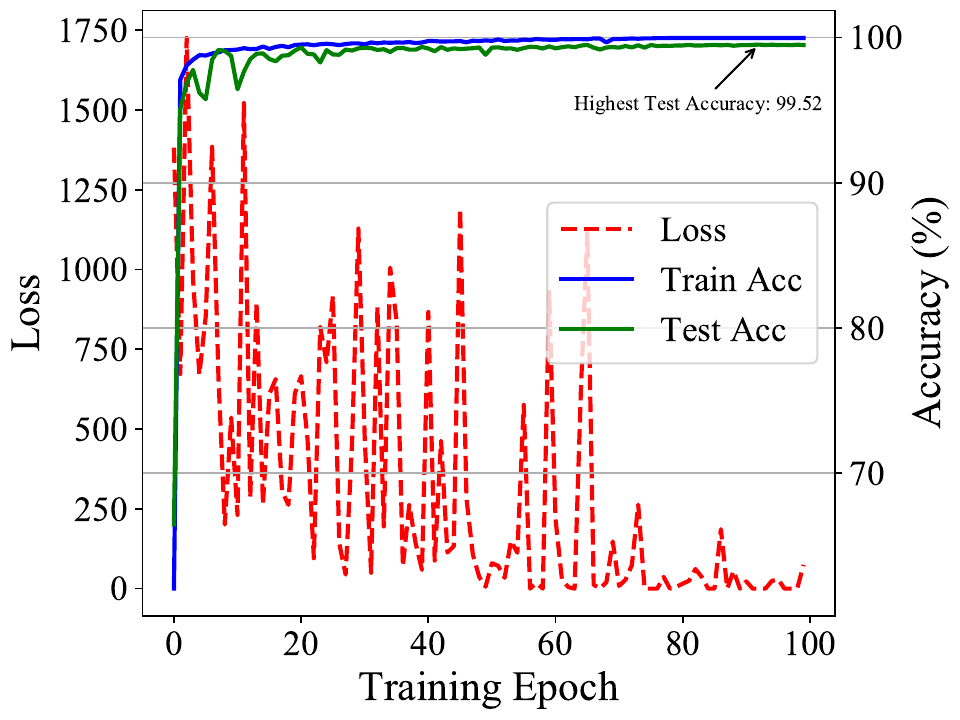}}
\subfigure[TRA CIFAR10]{
\includegraphics[width=0.48\columnwidth]{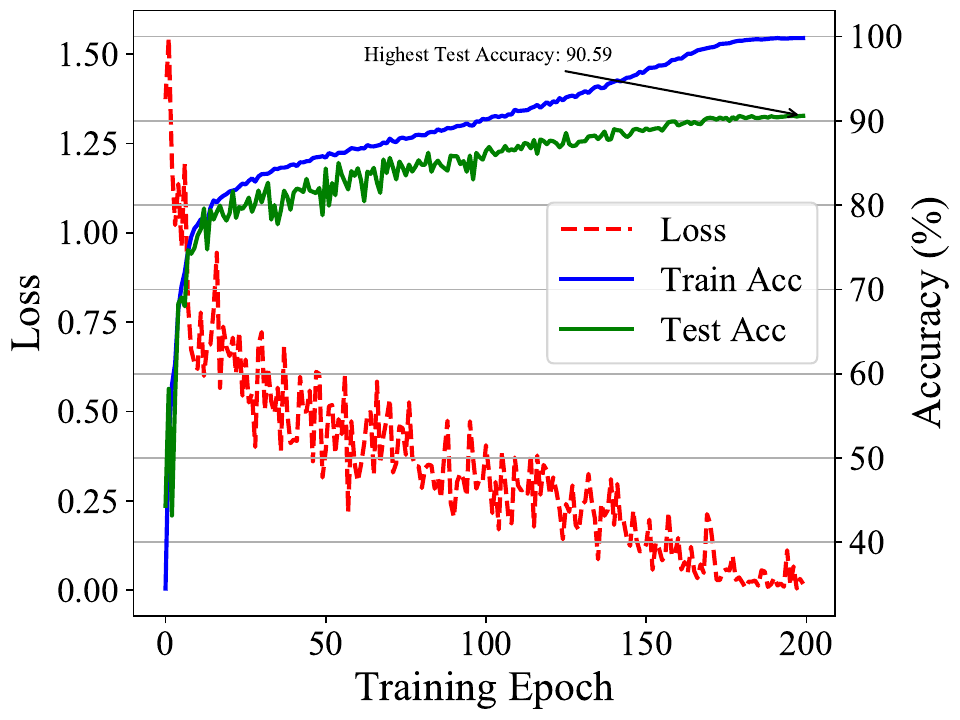}}
\subfigure[RBC CIFAR10]{
\includegraphics[width=0.48\columnwidth]{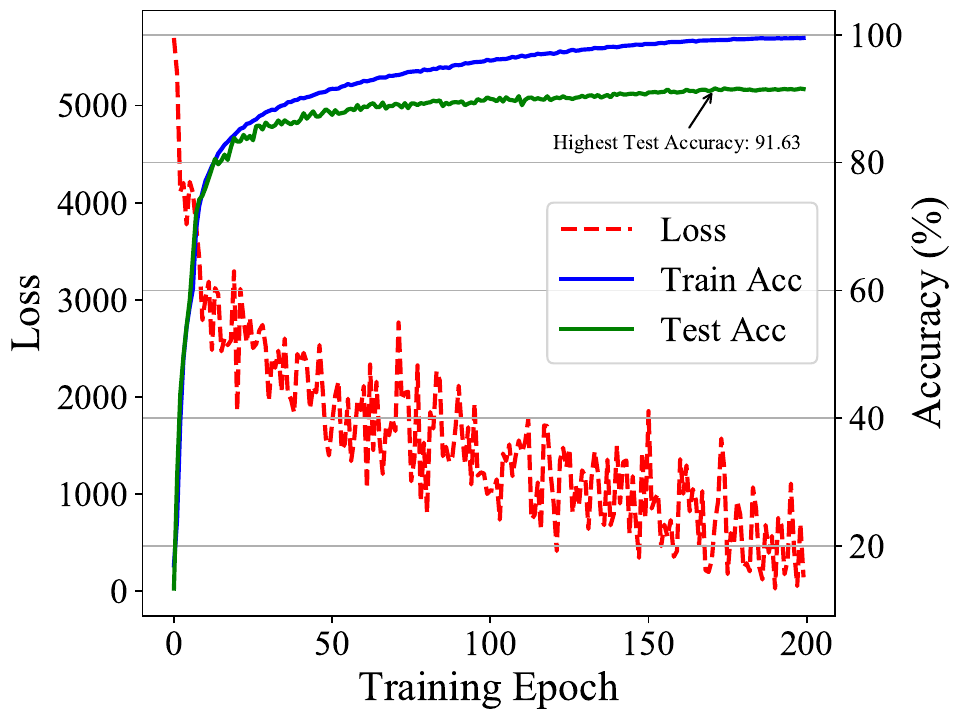}}
\subfigure[TRA CIFAR100]{
\centering
\includegraphics[width=0.48\columnwidth]{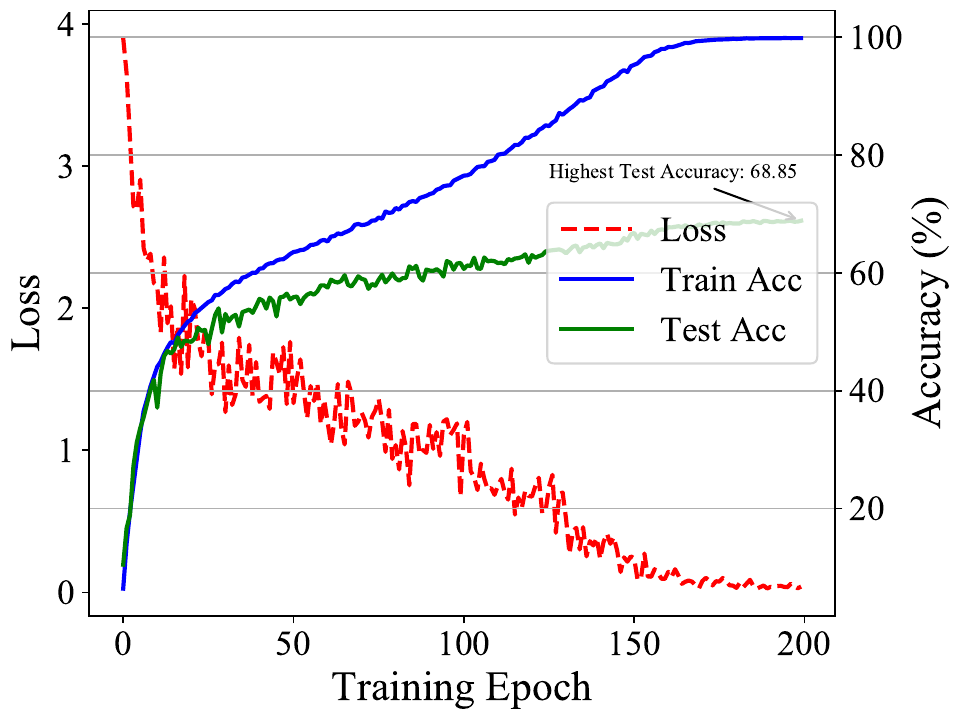}}
\subfigure[RBC CIFAR100]{
\centering
\includegraphics[width=0.48\columnwidth]{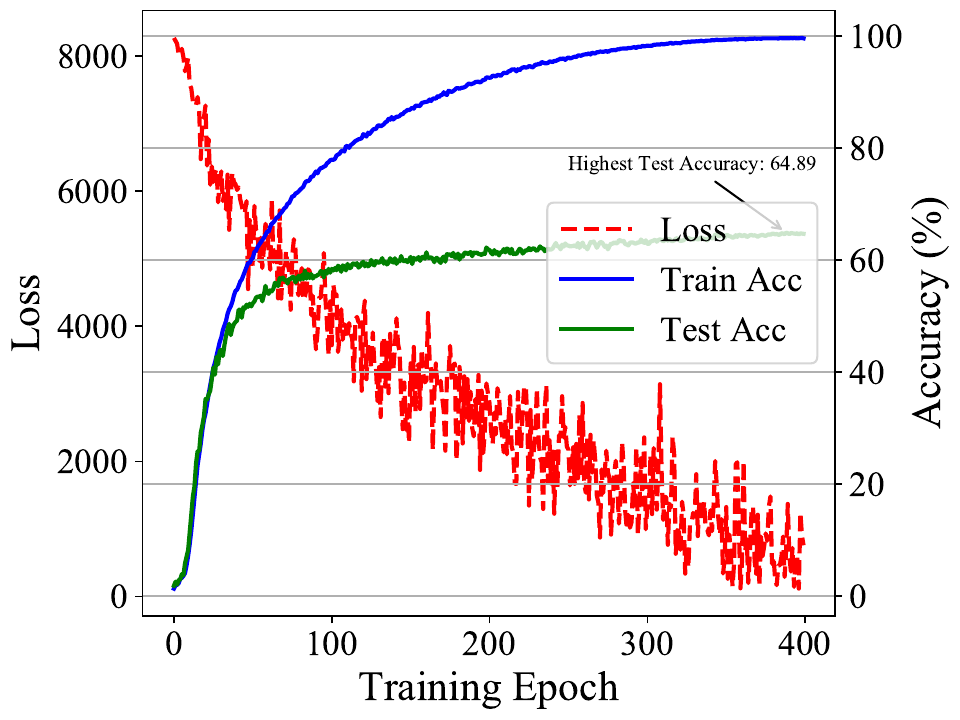}}
\caption{Comparison of convergence between the traditional and robust binary-label classifications. TRA and RBC represent the traditional and robust binary-label classification systems, respectively.}\label{convergence}
\end{figure}
In this subsection, we experiment to investigate the convergence of robust binary-label classification system. We compare the convergence of the traditional and binary-label classification systems on MNIST, CIFAR10, and CIFAR100. $B$ is set as 4 for MNIST and CIFAR10, while $B$ is set as 7 for CIFAR100. For the robust binary-label classification, the scaling factor $S$ is set as 350 for all datasets. In this paper, EfficientNet is EfficientNetB0 with 5.3M parameters \cite{efficientnet}, and ResNet is ResNet18 with 9.2M parameters \cite{resnet}.\\\\
As shown in Fig. \ref{convergence}, on MNIST and CIFAR10, robust binary-label classification system achieves higher test accuracy than the traditional classification system. While on CIFAR100, the binary-label classification system convergences more slowly than the traditional classification system. The reason is that the binary-label classification is more complex than the traditional classification. So the binary-label classification system needs a higher model capacity to fit the data distribution. The test accuracies of different classifications with variant models are similar, which demonstrates that when model capacity is sufficient, the robust binary-label and traditional classification systems generalize similarly. Moreover, that explains why the test accuracy of binary-label classification system is slightly lower than the traditional classification system in some cases.
\subsection{Robustness Evaluation for Different Classification Systems}\label{robsutness}
In this subsection, we experiment to investigate the robustness of different classification systems. We compare the robustness of the traditional classification, interval-label classification, and binary-label classifications with EfficientNet and ResNet on MNIST, CIFAR10, and CIFAR100. Since FGSM and PGD  are easy to control the attack intensity, we set three different attack intensities, namely \uppercase\expandafter{\romannumeral1}, \uppercase\expandafter{\romannumeral2}, and \uppercase\expandafter{\romannumeral3}. \uppercase\expandafter{\romannumeral1} represents the attack intensity is 0.1 for MNIST and 3/255 for CIFAR10 and CIFAR100. \uppercase\expandafter{\romannumeral2} represents the attack intensity is 0.2 for MNIST and 6/255 for CIFAR10 and CIFAR100. \uppercase\expandafter{\romannumeral3} represents the attack intensity is 0.3 for MNIST and 9/255 for CIFAR10 and CIFAR100. LEG in Table \ref{white-box} represents the accuracy of the legitimate examples. TRA, DeepBE, INT, and RBC represent the traditional classification, binary encoding classification, interval-label classification, and robust binary-label classification, respectively. For the binary-label classification, the $S$ is set 350, while for the interval-label classification, the interval label initial point $s_{0}$ is set as 0, interval label length $\beta$ is set as 16, and the gap length $\alpha$ is set as 4.\\\\
As shown in Table 1, our method achieves the highest accuracies in variant evaluation scenarios. Compared to the traditional and interval-label classification, our method can balance the adversarial robustness and accuracy better. We also note that the binary-label and interval-label classifications are insensitive to the variant in adversarial attack intensity, which is different from the traditional classification. The reason is that the traditional classification depends on the relative magnitude for the values of elements of the output vector, while the binary-label and interval-label classification depend on the actual magnitude for the values of elements in the outputs vector. Different adversarial attack intensities have a variant impact on the relative magnitude for the values of elements in the output vector, namely the logits. However, only when the adversarial attack intensity exceeds a specific threshold, the adversarial attack can change the sign of elements in the output vector or cause the values of elements in the output vector to vary from one interval to another.
\begin{table*}[htb]
\centering
\caption{Comparison of adversarial robustness and accuracy among different classification systems.}\label{white-box}
\setlength{\tabcolsep}{3mm}{
\begin{tabular}{r|c|c|ccc|ccc|c}
\toprule
\multirow{2}{*}{Data set}&\multirow{2}{*}{Model}&\multirow{2}{*}{Method}&\multicolumn{3}{c}{FGSM}&\multicolumn{3}{c|}{PGD}&\multirow{2}{*}{LEG}\\

\cmidrule{4-9}
&&&\uppercase\expandafter{\romannumeral1}&\uppercase\expandafter{\romannumeral2}&\uppercase\expandafter{\romannumeral3}&I&\uppercase\expandafter{\romannumeral2}&\uppercase\expandafter{\romannumeral3}&\\
\midrule
\multirow{8}{*}{MNIST}&\multirow{4}{*}{EfficientNetB0}&TRA&20.38&14.03&8.10&6.50&4.54&2.96&99.49\\
&&DeepBE&86.87&69.89&48.95&71.33&32.69&11.56&99.26\\
&&INT&99.12&99.12&99.13&99.04&98.71&\textbf{99.11}&99.18\\
&&RBC&\textbf{99.26}&\textbf{99.27}&\textbf{99.26}&\textbf{99.20}&\textbf{98.89}&97.90&\textbf{99.52}\\
\cmidrule{2-10}
&\multirow{4}{*}{ResNet18}&TRA&68.52&14.88&10.03&34.59&4.40&2.19&\textbf{99.65}\\
&&DeepBE&90.26&56.51&39.14&64.54&9.43&6.47&99.51\\
&&INT&\textbf{99.25}&99.21&\textbf{99.21}&99.16&\textbf{98.96}&\textbf{98.23}&99.25\\
&&RBC&99.21&\textbf{99.22}&\textbf{99.21}&\textbf{99.50}&95.19&74.55&99.55\\
\midrule
\multirow{8}{*}{CIFAR10}&\multirow{4}{*}{EfficientNetB0}&TRA&62.95&38.50&25.55&61.05&31.07&15.12&90.59\\
&&DeepBE&52.48&38.43&33.64&51.59&36.44&30.19&76.82\\
&&INT&80.66&80.37&80.46&80.48&80.19&80.30&83.94\\
&&RBC&\textbf{85.43}&\textbf{85.40}&\textbf{85.38}&\textbf{85.52}&\textbf{85.30}&\textbf{85.23}&\textbf{91.63}\\
\cmidrule{2-10}
&\multirow{4}{*}{ResNet18}&TRA&75.26&65.15&52.84&73.17&52.13&38.36&\textbf{93.67}\\
&&DeepBE&71.65&66.64&62.95&70.33&64.37&60.33&89.08\\
&&INT&86.47&86.55&86.56&86.40&86.55&86.47&87.54\\
&&RBC&\textbf{90.92}&\textbf{90.91}&\textbf{90.94}&\textbf{90.86}&\textbf{90.99}&\textbf{90.93}&93.55\\
\midrule
\multirow{8}{*}{CIFAR100}&\multirow{4}{*}{EfficientNetB0}&TRA&40.86&22.38&14.21&39.39&17.90&8.21&\textbf{68.85}\\
&&DeepBE&9.80&5.56&3.65&9.76&5.31&3.22&17.90\\
&&INT&6.10&6.12&6.21&6.04&6.06&6.09&8.24\\
&&RBC&\textbf{54.19}&\textbf{54.12}&\textbf{54.37}&\textbf{53.88}&\textbf{53.76}&\textbf{53.75}&64.68\\
\cmidrule{2-10}
&\multirow{4}{*}{ResNet18}&TRA&46.52&31.87&24.06&42.94&23.16&12.37&\textbf{72.05}\\
&&DeepBE&16.68&11.06&8.76&16.40&10.32&7.98&27.99\\
&&INT&17.59&17.61&17.65&17.43&17.52&17.62&20.13\\
&&RBC&\textbf{58.33}&\textbf{58.35}&\textbf{58.45}&\textbf{58.35}&\textbf{58.26}&\textbf{57.94}&69.09\\

\bottomrule
\end{tabular}}
\end{table*}

\begin{table*}[htb]
\centering
\caption{Comparison of adversarial robustness and accuracy between traditional and robust bianry-label classification systems with Madry Adversarial Training.}\label{r_at}
\setlength{\tabcolsep}{3mm}{
\begin{tabular}{r|c|c|ccc|ccc|c}
\toprule
\multirow{2}{*}{Data set}&\multirow{2}{*}{Model}&\multirow{2}{*}{Method}&\multicolumn{3}{c}{FGSM}&\multicolumn{3}{c|}{PGD}&\multirow{2}{*}{LEG}\\
\cmidrule{4-9}
&&&\uppercase\expandafter{\romannumeral1}&\uppercase\expandafter{\romannumeral2}&\uppercase\expandafter{\romannumeral3}&I&\uppercase\expandafter{\romannumeral2}&\uppercase\expandafter{\romannumeral3}&\\
\midrule
\multirow{6}{*}{MNIST}&\multirow{3}{*}{EfficientNetB0}
&TRA+AT&98.48$\uparrow$&96.89$\uparrow$&88.65$\uparrow$&98.37$\uparrow$&80.51$\uparrow$&10.36$\uparrow$&99.44$\downarrow$\\
&&RBC+AT&\textbf{99.21}$\downarrow$&{99.20}$\downarrow$&{99.20}$\downarrow$&{99.19}$\downarrow$&{99.18}$\uparrow$&\textbf{99.14}$\uparrow$&\textbf{99.54}$\uparrow$\\
&&RBC+GN&\textbf{99.34}$\uparrow$&\textbf{99.33}$\uparrow$&\textbf{99.35}$\uparrow$&\textbf{99.36}$\uparrow$&\textbf{99.21}$\uparrow$&{98.78}$\uparrow$&{99.44}$\downarrow$\\
\cmidrule{2-10}
&\multirow{3}{*}{ResNet18}
&TRA+AT&98.97$\uparrow$&98.00$\uparrow$&85.03$\uparrow$&\textbf{98.98}$\uparrow$&94.87$\uparrow$&30.25$\uparrow$&\textbf{99.57}$\downarrow$\\
&&RBC+AT&{99.04}$\downarrow$&{98.87}$\downarrow$&{98.88}$\downarrow$&98.96$\downarrow$&{98.89}$\uparrow$&\textbf{98.77}$\uparrow$&99.53$\downarrow$\\
&&RBC+GN&\textbf{99.30}$\uparrow$&\textbf{99.30}$\uparrow$&\textbf{99.30}$\uparrow$&\textbf{99.33}$\downarrow$&\textbf{99.00}$\uparrow$&{94.28}$\uparrow$&{99.56}$\uparrow$\\
\midrule
\multirow{6}{*}{CIFAR10}&\multirow{3}{*}{EfficientNetB0}&TRA+AT&80.33$\uparrow$&69.25$\uparrow$&58.78$\uparrow$&80.30$\uparrow$&69.41$\uparrow$&57.81$\uparrow$&88.58$\downarrow$\\
&&RBC+AT&{83.31}$\downarrow$&{79.81}$\downarrow$&{79.22}$\downarrow$&{83.25}$\downarrow$&{79.60}$\downarrow$&{79.18}$\downarrow$&{91.30}$\downarrow$\\
&&RBC+GN&\textbf{88.43}$\uparrow$&\textbf{88.48}$\uparrow$&\textbf{88.49}$\uparrow$&\textbf{88.46}$\uparrow$&\textbf{88.53}$\uparrow$&\textbf{88.27}$\uparrow$&\textbf{92.09}$\uparrow$\\
\cmidrule{2-10}
&\multirow{3}{*}{ResNet18}&TRA+AT&86.81$\uparrow$&80.11$\uparrow$&72.96$\uparrow$&84.74$\uparrow$&79.66$\uparrow$&71.41$\uparrow$&92.63$\downarrow$\\
&&RBC+AT&{86.94}$\downarrow$&{86.10}$\downarrow$&{86.03}$\downarrow$&{86.92}$\downarrow$&{86.07}$\downarrow$&{85.95}$\downarrow$&{93.29}$\downarrow$\\
&&RBC+GN&\textbf{91.75}$\uparrow$&\textbf{91.77}$\uparrow$&\textbf{91.78}$\uparrow$&\textbf{91.72}$\uparrow$&\textbf{91.70}$\uparrow$&\textbf{91.52}$\uparrow$&\textbf{93.78}$\uparrow$\\
\midrule
\multirow{6}{*}{CIFAR100}&\multirow{3}{*}{EfficientNetB0}&TRA+AT&{53.66}$\uparrow$&43.16$\uparrow$&34.03$\uparrow$&53.63$\uparrow$&42.78$\uparrow$&33.14$\uparrow$&\textbf{64.70}$\downarrow$\\
&&RBC+AT&{50.91}$\downarrow$&{49.60}$\downarrow$&{49.21}$\downarrow$&{50.52}$\downarrow$&{49.03}$\downarrow$&{48.86}$\downarrow$&63.36$\downarrow$\\
&&RBC+GN&\textbf{57.30}$\uparrow$&\textbf{57.18}$\uparrow$&\textbf{57.44}$\uparrow$&\textbf{56.85}$\uparrow$&\textbf{56.75}$\uparrow$&\textbf{56.67}$\uparrow$&64.30$\downarrow$\\
\cmidrule{2-10}
&\multirow{3}{*}{ResNet18}&TRA+AT&{59.07}$\uparrow$&50.38$\uparrow$&42.32$\uparrow$&58.87$\uparrow$&49.69$\uparrow$&40.87$\uparrow$&\textbf{68.96}$\downarrow$\\
&&RBC+AT&52.81$\downarrow$&{51.17}$\downarrow$&{51.08}$\downarrow$&{52.54}$\downarrow$&{51.18}$\downarrow$&{51.00}$\downarrow$&66.94$\downarrow$\\
&&RBC+GN&\textbf{61.22}$\uparrow$&\textbf{61.29}$\uparrow$&\textbf{61.40}$\uparrow$&\textbf{61.24}$\uparrow$&\textbf{61.09}$\uparrow$&\textbf{60.95}$\uparrow$&68.56$\downarrow$\\
\bottomrule
\end{tabular}}
\end{table*}
\noindent We also experiment to investigate the robustness of our method combining with the adversarial training or retraining with Gaussian Noise perturbed examples. The adversarial training method in this paper is Madry adversarial training \cite{pgdat}. The intensity of adversarial examples and Gaussian Noise for the adversarial training and retraining is set as \uppercase\expandafter{\romannumeral1}, as mentioned above. In Table \ref{r_at}, the TRA+AT and RBC+AT represent the traditional classification with Madry adversarial training and robust binary-label classification with Madry adversarial training, respectively. RBC+GN represents the robust binary-label classification retrained with Gaussian Noise perturbed examples. $\uparrow$ represents that corresponding retraining method can improve the adversarial robustness, while $\downarrow$ represents the retraining suppresses the adversarial robustness.\\\\
As shown in Table \ref{r_at}, the robust binary-label classification is still more robust than the traditional classification, even with the adversarial training. We note that the adversarial training with a specific adversarial attack intensity can not endow the deep model robustness against variant-intensity attacks, especially when the attack intensity is bigger than that in the adversarial training. However, the variant of the attack intensity has a limited impact on the robustness of the RBC. Different from the traditional classification, we note that adversarial training can not constantly improve the robustness of RBC, but retraining with Gaussian Noise perturbed examples can. Here we define a stability metric of robust accuracy.
\begin{equation}
\mathcal{R}=\frac{acc_{a}}{acc_{t}},
\end{equation} 
where $acc_{a}$ and $acc_{t}$ are robust accuracy and test accuracy, when adopting specific defense strategy. $\mathcal{R}$ can be employed to evaluate the sufficiency of defense method. For instance, if the $\mathcal{R}$ is smaller than 1 for the adversarial training, we may need to increase the training epochs or the attack intensity. According to Table \ref{r_at}, adversarial training improves the robustness of RBC, when $\mathcal{R}$ is abnormally small. However, when $\mathcal{R}$ is large enough, adversarial training will inhibit the robustness of RBC.\\\\
\begin{figure*}[htb]
\centering
\subfigure[TRA]{
\includegraphics[width=0.51\columnwidth]{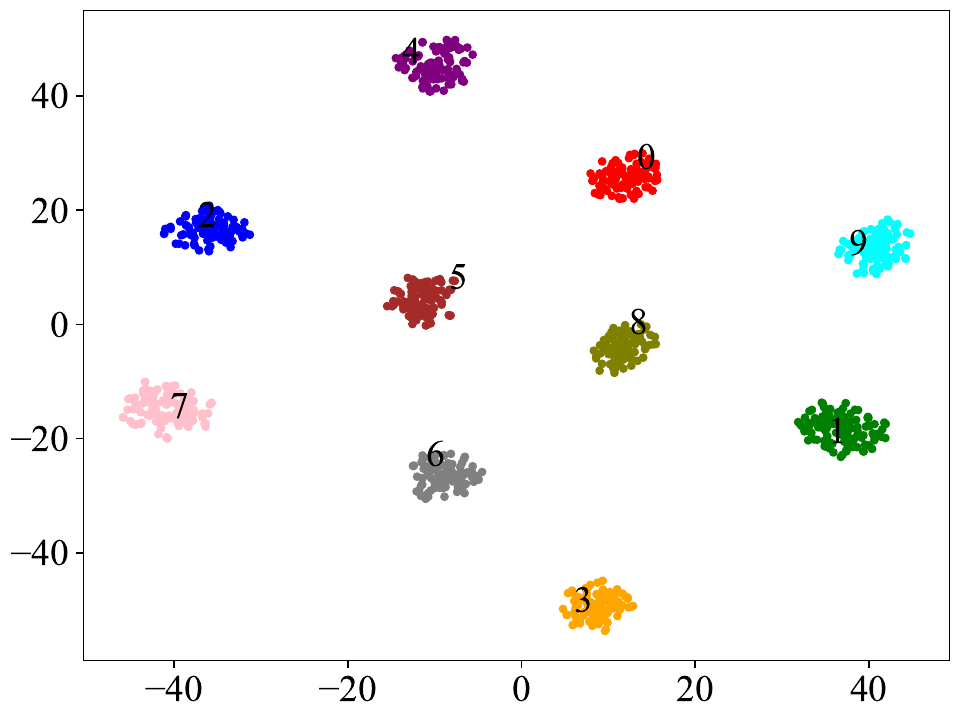}}
\subfigure[DeepBE]{
\includegraphics[width=0.51\columnwidth]{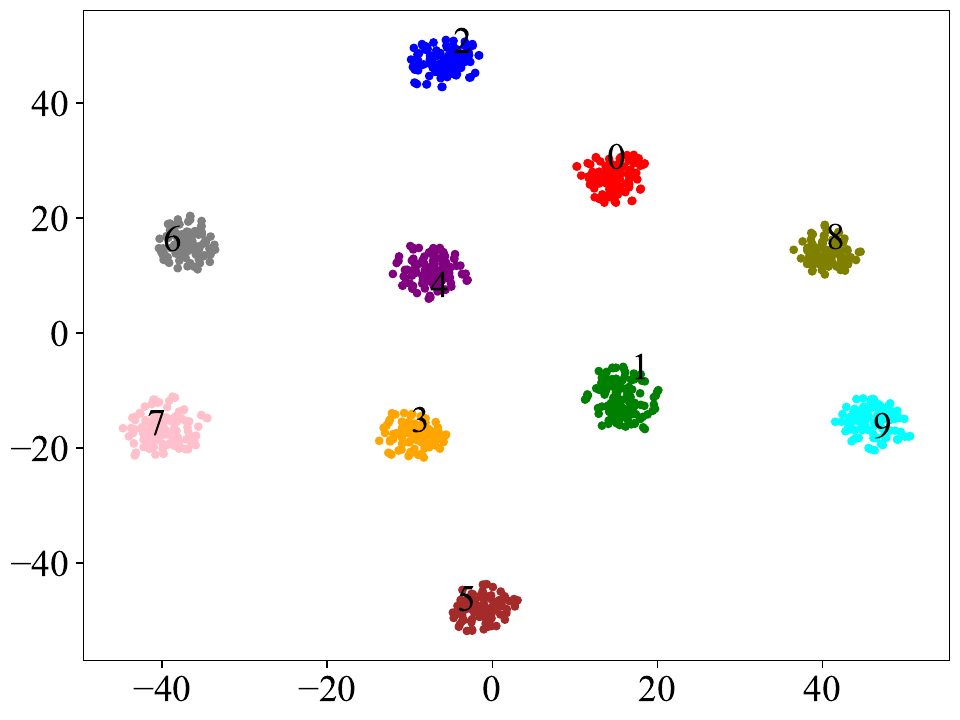}}
\subfigure[RBC]{
\includegraphics[width=0.51\columnwidth]{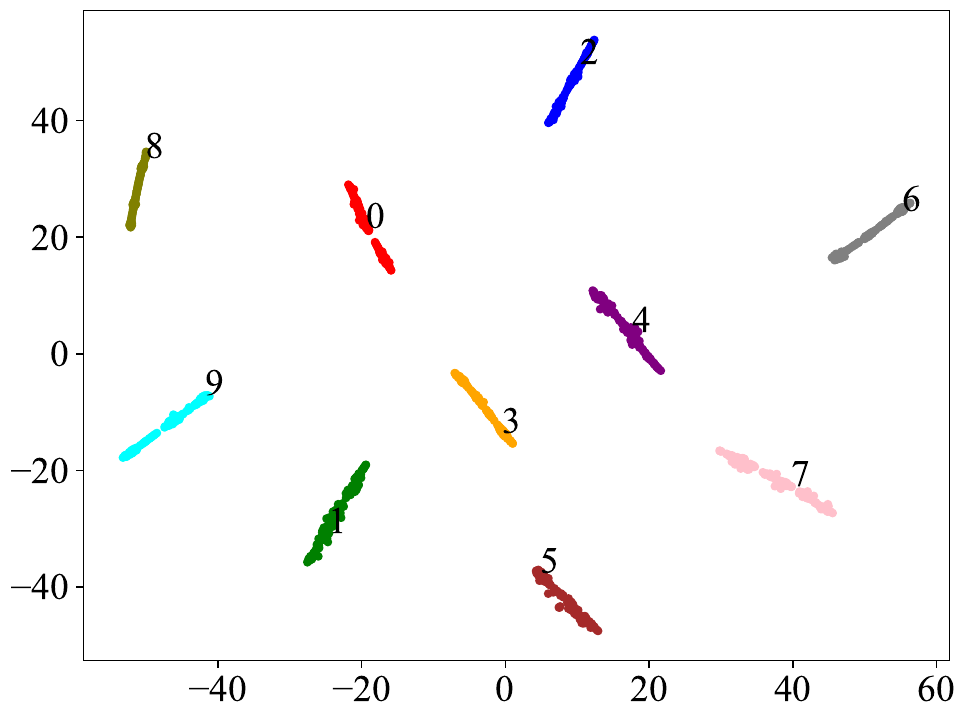}}
\subfigure[INT]{
\includegraphics[width=0.51\columnwidth]{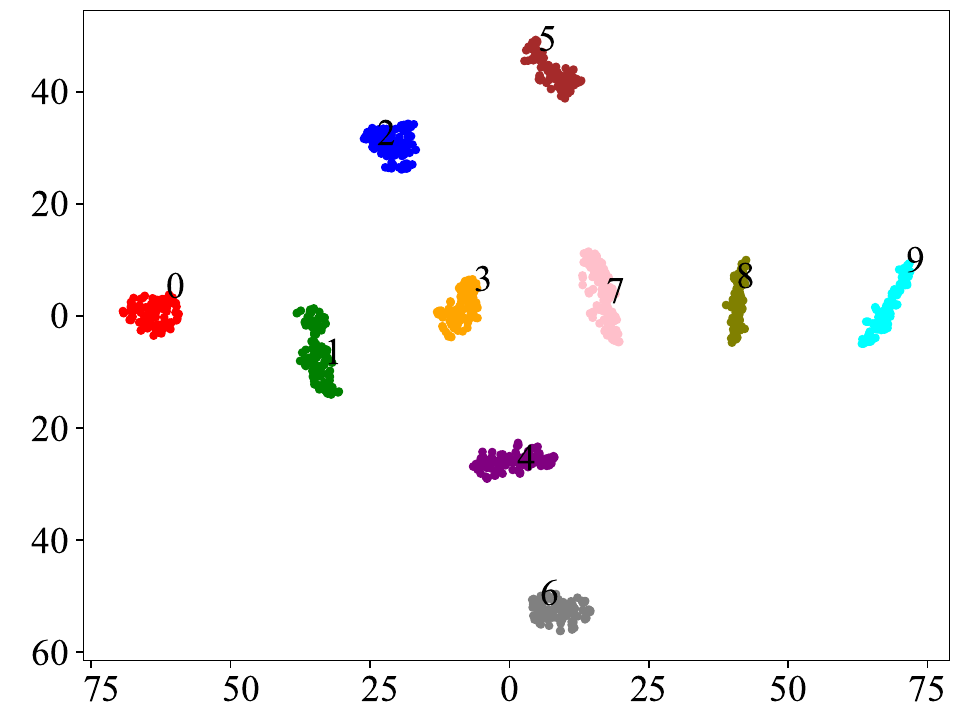}}
\caption{The t-SNE visualization of features in penultimate layer of EfficientNetB0 for traditional classification (TRA), DeepBE, robust binary-label classification (RBC), and interval-label classification (INT) on MNIST. The `red', `green', `blue',  `orange',  `purple', `brown', `gray', `pink', `olive', and `cyan' colorful points represent examples of `0', `1', `2', `3', `4', `5', `6', `7', `8', and `9' category, respectively.}\label{tsne}
\end{figure*}
To further intestigate robustness of variant classification systems, we also experiment to visualize features with t-SNE \cite{t-sne}. Since the EfficientNetB0 has been employed as the backbone model for many learning tasks, we apply t-SNE visualization to its penultimate layer features. We randomly select 1,000 examples from the MNIST for the visualization. The scaling factor $S$ is set as 350 for the robust binary-label classification.\\\\
As shown in Fig. \ref{tsne}, EfficientNetB0 can learn a reasonable feature representation for the traditional, binary-label, and interval label classification. There is even no intersection for essential features of different examples belong to different categories. Moreover, we note that features of examples from the same category are distributed in long and narrow areas for the robust binary-label and interval-label classification. The reason is that robust binary-label and interval label classification depends on actual values of the outputs, and the loss function constraints their outputs into some specific regions. We also note that the distances between features of different categories for robust binary-label and interval-label classification are more significant than that of traditional classification or DeepBE. Increasing the margin between the boundaries of different categories can improve the robustness of deep learning systems \cite{tilting}. That is the reason why robust binary-label and interval-label classifications are more robust than the traditional classification.\\\\
Fig. \ref{tsne} also can explain different impacts of adversarial training on the traditional classification and robust binary classificaiton. The feature distribution manifold for traditional classification approximates a hypersphere. The adversarial example is near to the decision boundary, which can make the decision boundary more smooth and decrease the dimension of adversarial space \cite{thespace}. However, features of RBC distribute in a long and narrow regions. Moreover, RBC depends on actual values of elements in output vector. The adversarial training will extend the feature manifold and decrease the margin between different categories. The feature manifold extension has a limited impact on traditional classification since it depends on the relative values of elements in the output vector.
\subsection{Impact of $S$ on Accuracy and Robustness of Robust Binary-label Classification System}
\begin{figure}[htb]
\centering
\includegraphics[width=0.85\columnwidth]{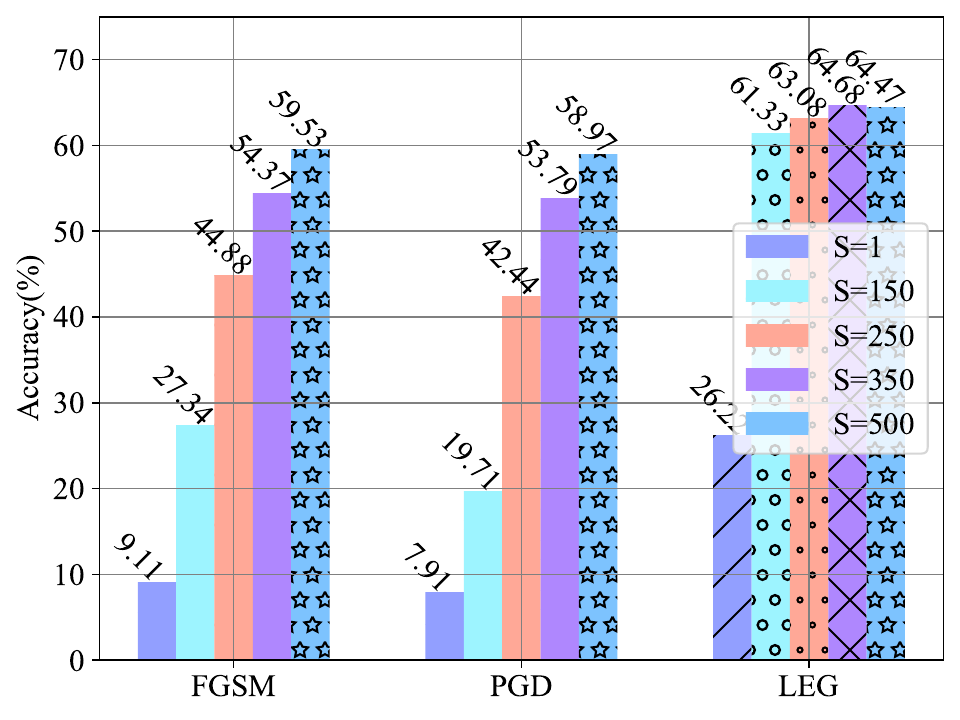}
\caption{Comparison of accuracy and adversarial robustness between robust binary-label classifications with variant $S$.}\label{s}
\end{figure}
In this subsection, we experiment to investigate the impact of $S$\ in Eq. \ref{loss} on the accuracy and robustness of the robust binary-label classification system. In detail, we set variant $S$ for RBC to check its accuracy and robustness on CIFAR100. The deep model to implement RBC is the EfficientNetB0. Moreover, the FGSM and PGD-5 are adopted, and the adversarial intensity is set as 9/255.\\\\
As shown in Fig. \ref{s}, with increasing $S$, the robust binary-label classification achieves higher test accuracy and robustness. The reason is that $S$ determines the margin between different categories. Bigger $S$ endows RBC with a bigger margin between different categories. Furthermover, bigger margin endows the robust binary-label classification to generalize better. However, we also note that when $S$ exceeds a threshold, the generalization of the robust binary-label classification improves slowly. Moreover, we note that the increasing $S$ will slow down the convergence. When $S=350$, the robust binary-label classification achieves reasonable convergence. According to Fig. \ref{s}, we conclude that the accuracy and robustness for the robust binary-label classification are consistent, which is different from the traditional classification \cite{odds}.
\subsection{Investigation of Adversarial Transferability for Different Classification Systems}
\begin{figure}
\centering
\subfigure[TRA]{
\includegraphics[width=0.45\columnwidth]{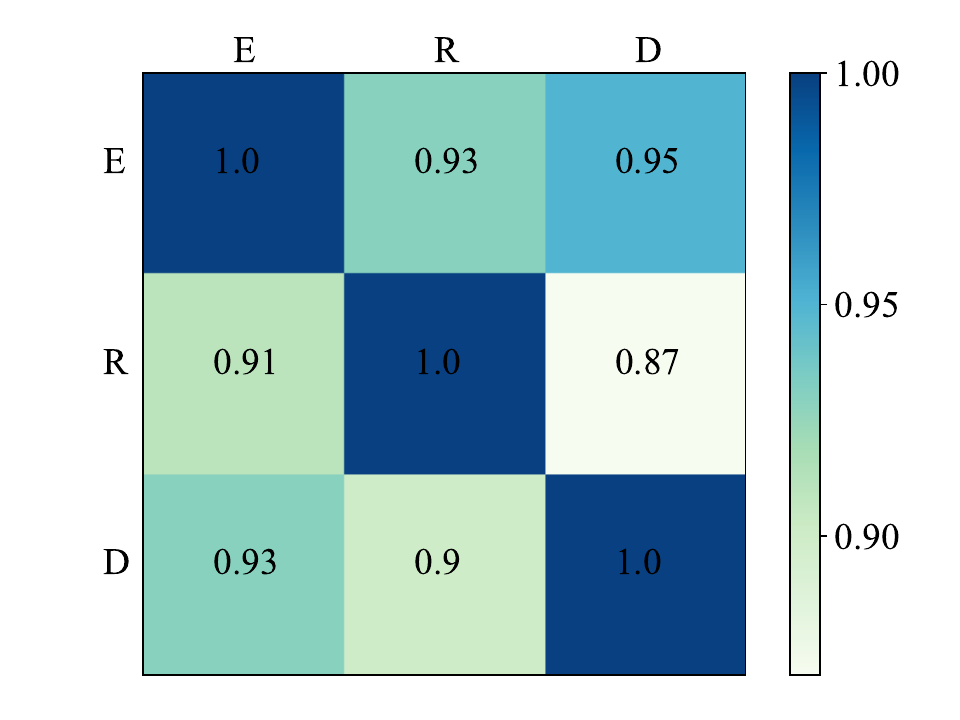}}
\subfigure[INT]{
\includegraphics[width=0.45\columnwidth]{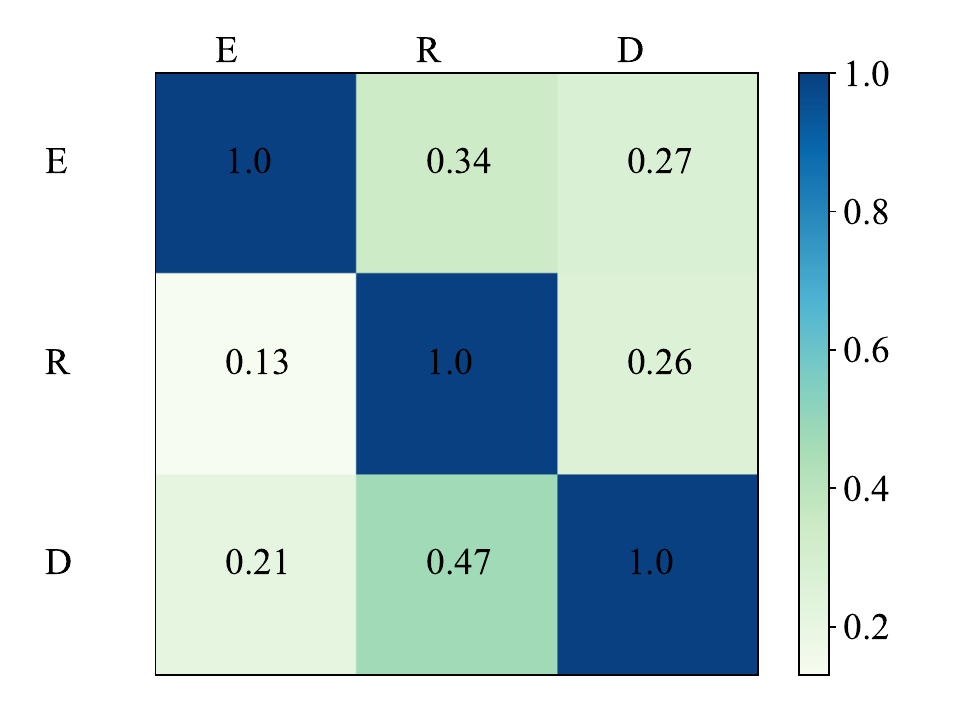}}
\subfigure[RBC $S$\label{s tran}]{
\includegraphics[width=0.45\columnwidth]{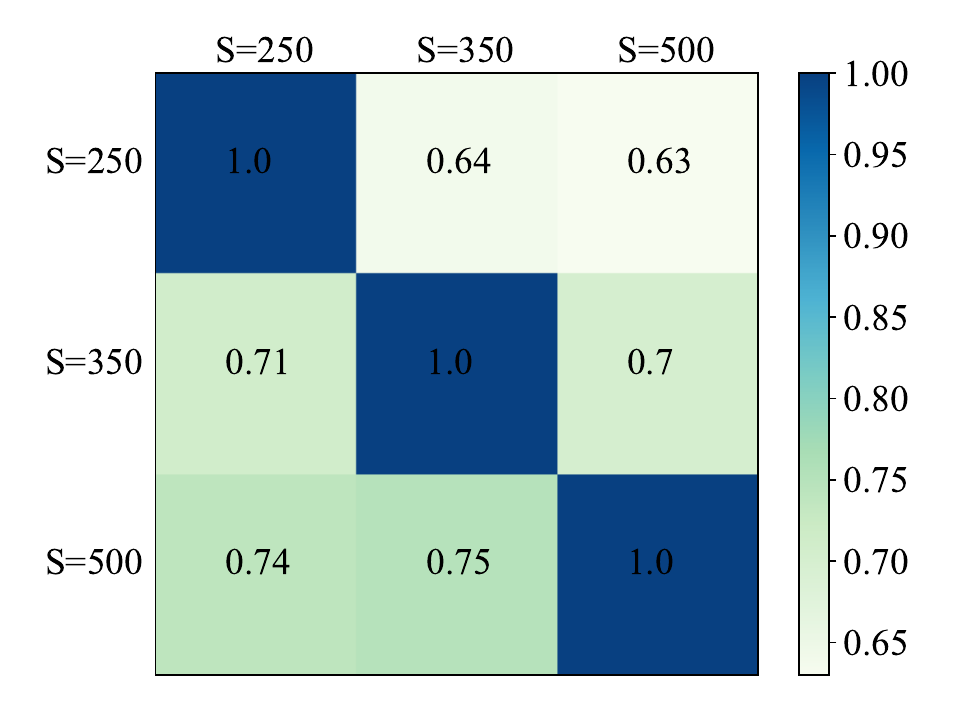}}
\subfigure[RBC ARC\label{arc tran}]{
\includegraphics[width=0.45\columnwidth]{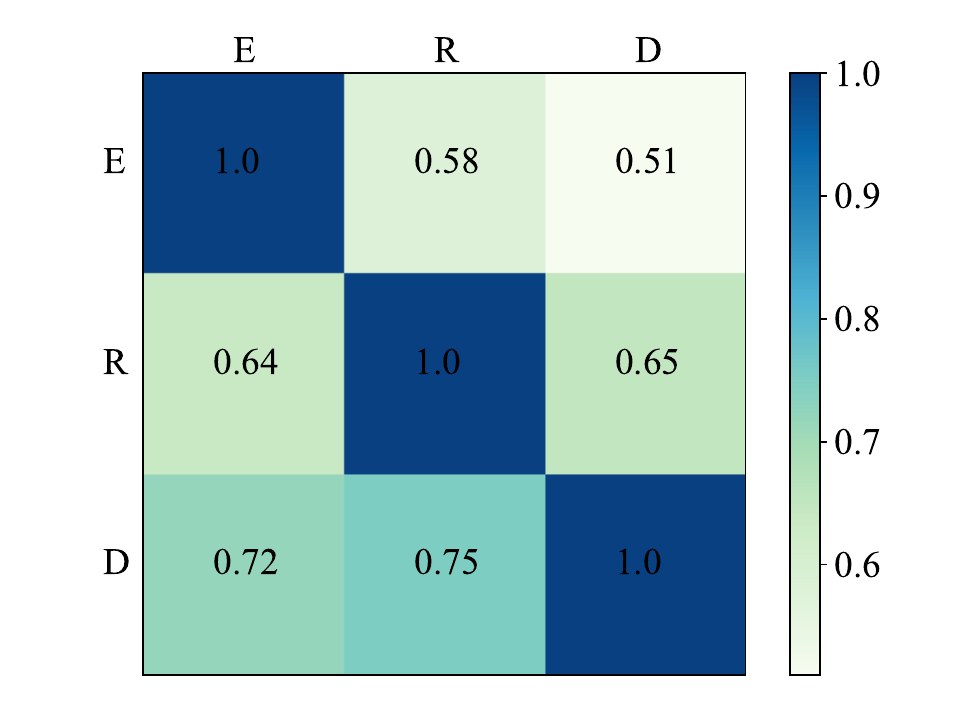}}
\caption{The adversarial transferability of different classification systems. The values in the figure represent the success attack rate. E, R, and D represent EfficientNetB0, ResNet18, and DenseNet121, respectively. Models in the row are threat models, while models in the column are victim models. Fig. \ref{s tran} investigates the adversarial transfer rate for RBC with variant $S$, and the adopted deep model is EfficientNetB0. Fig. \ref{arc tran} investigates the adversarial transfer rate for RBC implemented by different deep models, and $S$ is set as 350.}\label{tran}
\end{figure}
In this subsection, we experiment to investigate the adversarial transferability between different models for the traditional classification, interval-label, and robust binary-label classification systems. In detail, CIFAR100 is employed evaluation dataset to investigate the adversarial transferability. The adopted adversarial attack method is PGD-5, and the adversarial attack intensity is set as 9/255. In Fig. \ref{tran}, the model in the row is the threat model, while the model in the column is the victim model. The threat model crafts adversarial examples to attack victim models. In order to eliminate the impact of test accuracy of the threat model, we remove examples that are misclassified by the threat model from the testing set of CIFAR100.\\\\
As shown in Fig. \ref{tran}, the adversarial transfer rate of the traditional classification is higher than that of the robust binary-label and interval-label classification. We also attribute it to the reason that the traditional classification depends on the relative magnitude for the values of elements in the output vector, while the binary-label and interval-label classification depend on the actual magnitude for the values of elements in the outputs vector. The adversarial attack can change the relative magnitude for element values but different magnitude of actual values in the output vector \cite{atn}.
\section{Discussion}
The DeepBE can be viewed as a special case of RBC. The loss of DeepBE \cite{deepbe} can be reformulated as $\|r(\mathbf{0}-(\mathcal{B}(x)-0.5)(2\bm{b}-1))\|_{2}^{2}$. So the DeepBE can be viewed as RBC with $S=0$. Here $\mathcal{B}$ and $r$ the binary-label classifier and ReLU function, respectively. $x$ is an example, and $\bm{b}$ is its corresponding binary label. The relationship between our method and regularization methods, such as adversarial training, is parallel. They focus on different elements of the deep learning system. As introduced above, our method with adversarial training can further improve the robustness of the deep learning system in some cases.\\\\ 
The interval-label classification and RBC adopt different label encoding strategies to redefine the classification task. Actually, with an increasing numerical system, we can adopt fewer bits to represent a number. For instance, decimal number 9 is only a bit in decimal encoding, while it needs four bit in binary encoding. Interval-label classification can be viewed as adopting a larger numerical system than binary-label classification to encoding the label. They both mark the input depending on the actual values of elements in the output vector. 
\section{Conclusion}
In this paper, we improve the robustness of deep learning systems from both the learning task and deep model prospects. The interval-label classification can reduce the adversarial transferability of adversarial examples. Unlike the traditional classification, the robustness of robust binary-label classification is no more at odds with the accuracy. The scaling factor of the robust binary-label classification constrains the minima boundary margin. Moreover, the experimental results demonstrate that adversarial training does not constantly improve the adversarial robustness of the deep learning system, demonstrating that the learning task impacts the robustness of deep learning systems.
\bibliography{Formatting-Instructions-LaTeX-2022}


\bigskip

\end{document}